\documentclass[letterpaper]{article} 
\usepackage{aaai25}  
\usepackage{times}  
\usepackage{helvet}  
\usepackage{courier}  
\usepackage[hyphens]{url}  
\usepackage{graphicx} 
\usepackage{natbib}  
\usepackage{caption} 
\usepackage{mathtools}
\usepackage{amssymb}
\usepackage{multirow} 
\usepackage{arydshln} 
\usepackage[table]{xcolor} 
\usepackage{array} 
\usepackage{booktabs} 
\usepackage{makecell} 

\title{Joint Knowledge Editing for Information Enrichment and Probability Promotion}
\author{
    Wenhang Shi\textsuperscript{\rm 1}, 
    Yiren Chen\textsuperscript{\rm 2},
    Shuqing Bian\textsuperscript{\rm 3},
    Xinyi Zhang\textsuperscript{\rm 1}\thanks{Corresponding Author},
    Zhe Zhao\textsuperscript{\rm 3}\textsuperscript{\raisebox{-0.5ex} {\normalfont*}},
    Pengfei Hu\textsuperscript{\rm 3},
    Wei Lu\textsuperscript{\rm 1},
    Xiaoyong Du\textsuperscript{\rm 1}
}
\affiliations{
    \textsuperscript{\rm 1} Renmin University of China, \textsuperscript{\rm 2}Peking University, \textsuperscript{\rm 3}Tencent\\

    \{wenhangshi, xinyizhang.info, lu-wei, duyong\}@ruc.edu.cn, yrchen92@pku.edu.cn, shuqingbian@gmail.com, \\ \{nlpzhezhao, alanpfhu\}@tencent.com

}

\begin{document}

\maketitle
\begin{abstract}
Knowledge stored in large language models requires timely updates to reflect the dynamic nature of real-world information.
To update the knowledge, most knowledge editing methods focus on the low layers, since recent probes into the knowledge recall process reveal that the answer information is enriched in low layers.
However, these probes only and could only reveal critical recall stages for the original answers, while the goal of editing is to rectify model's prediction for the target answers.
This inconsistency indicates that both the probe approaches and the associated editing methods are deficient.
To mitigate the inconsistency and identify critical editing regions, we propose a contrast-based probe approach, and locate two crucial stages where the model behavior diverges between the original and target answers: \textbf{Information Enrichment} in low layers and \textbf{Probability Promotion} in high layers.
Building upon the insights, we develop the \textbf{J}oint knowledge \textbf{E}diting for information \textbf{E}nrichment and probability \textbf{P}romotion (JEEP) method, which jointly edits both the low and high layers to modify the two critical recall stages. Considering the mutual interference and growing forgetting due to dual modifications, JEEP is designed to ensure that updates to distinct regions share the same objectives and are complementary. 
We rigorously evaluate JEEP by editing up to thousands of facts on various models, \textit{i.e.}, GPT-J (6B) and LLaMA (7B), and addressing diverse editing objectives, \textit{i.e.}, adding factual and counterfactual knowledge. In all tested scenarios, JEEP achieves best performances, validating the effectiveness of the revealings of our probe approach and the designs of our editing method. Our code and data are available at \url{https://github.com/Eric8932/JEEP}.

\end{abstract}

\begin{figure}[t]
\centering
\includegraphics[scale = 0.42]{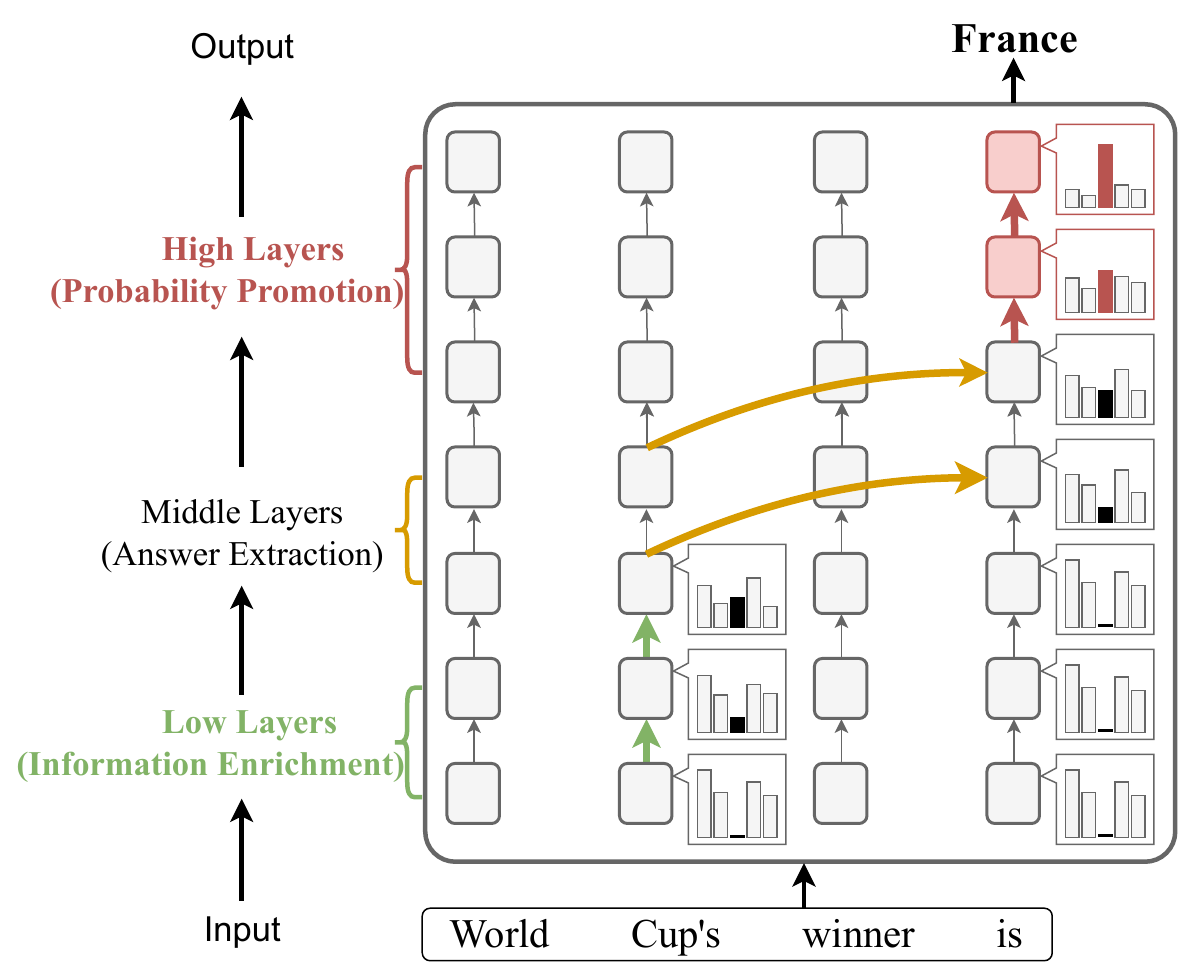}
\caption{Using probability as information indicator, we directly observe the original answer's information flow within the model. By further contrasting information flow of the original and target answers, we identify two critical recall stages for knowledge editing: Information Enrichment in low layers and Probability Promotion in high layers.} 
\label{pic1}
\end{figure}

\section{Introduction}
Large Language Models (LLMs) are renowned for their extensive knowledge storage, addressing queries by recalling the encoded knowledge \cite{petroni2019language,touvron2023llama}. However, their original knowledge might be incorrect or outdated due to the swift pace of global events, demanding timely updates to this stored information \cite{jang2022temporalwiki}. For instance, the answer ``France" to the query ``World Cup's winner is" remains valid until year 2022, but now ``Argentina" is the correct response. Common methods like retraining model with revised corpora is prohibitively costly, while continually fine-tuning on the corrected dataset causes catastrophic forgetting \cite{mccloskey1989catastrophic,kirkpatrick2017overcoming,DBLP:conf/ecai/ShiCZLYD23}. Consequently, there is a growing interest in \textit{Knowledge Edit}, whose goal is to update the model's original knowledge to the target knowledge both efficiently and accurately \cite{de2021editing,mitchell2022memory}.

Knowledge editing methods can be categorized based on whether they modify model parameters \cite{yao2023editing}. Weight-preserved methods incorporate additional structures to handle each editing requirement, facing scalability issues as the number of edits increases. \cite{huang2023transformer,hartvigsen2024aging,yu2024melo}. 
Therefore, we focus on weight-modified methods, which could accommodate considerable updates in a single operation, and include two distinct methods \cite{meng2022mass,tan2023massive,li2024pmet}. Meta-learning methods train hyper-networks for generating updated parameters, suffering from poor generalization when edits increase \cite{mitchell2021fast, tan2023massive}. Locate-then-edit methods locate primary storage locations of knowledge, by probing into model's recall process of the original answers, then edit the specific locations \cite{dai2021knowledge, meng2022locating}. 
Recent explorations show that the output answer processing in LLMs involves two critical phases: information enrichment in low layers and answer extraction in middle layers \cite{meng2022locating,geva2023dissecting}. Building on the insights, locate-then-edit methods all focus updates on the low layers \cite{meng2022locating,meng2022mass, li2024pmet}. 

Despite the demonstrated capability of existing locate-then-edit methods for large-scale knowledge editing \cite{meng2022mass, li2024pmet}, their effectiveness is hampered by the inconsistency between the probe approaches for original answers and the editing goals for target answers. 
Previous probes assess the impact of ablating specific modules by comparing the probabilities of the original answers in model's predictions before and after such interventions. This reliance on final probabilities limits their applicability to the target answers, which have low probabilities in the predictions. However, the stages vital for effective knowledge editing do not merely align with those critical for recalling original answers, but are distinctively associated with the stages where the model's behavior diverges between the original and target answers. To rectify the inconsistency, we propose a contrast-based probe approach. It observes the answer's probability across different representations as in Fig. \ref{pic1}, which depicts the original answer's information flow through each layer \cite{logitlens}. By contrasting information changes of the original and target answers, we identify two critical stages for knowledge editing: \textbf{Information Enrichment} in low layers and \textbf{Probability Promotion} in high layers. 
The low layers enrich answer-related information to the representations, and the high layers promote answer's probability to make it the final output.

Building on these insights, we develop the JEEP method, which strategically targets both low and high layers for a holistic alternation of knowledge recall process. It jointly edits layers responsible for information enrichment and probability promotion stages. However, modifications to different model regions can interact in complex ways, potentially leading to conflicting outcomes. To ensure the updates are not only cohesively integrated but also complementary, JEEP first synergizes the optimization objectives of both updates. It then adaptively adjusts the update degree for different layers based on the information changes required at each recall stage, thereby effectively altering model's predictions with minimal parameter changes.
To validate our method, we conduct extensive experiments involving edits ranging from 1 to 10,000 across various model architectures, including GPT-J (6B) and LLaMA (7B) \cite{wang2021gpt,touvron2023llama}, and datasets such as zsRE and Multi-COUNTERFACT \cite{levy2017zero,meng2022mass}. In all tested scenarios, JEEP consistently delivers the optimal performances, confirming the effectiveness of our methodological designs and validating our probe approach to identify critical editing stages.

The main contributions of our study are threefold:
$\left ( 1 \right ) $ We identify the inconsistency between existing probes for the original answers and the editing goals for the target answers. By introducing a contrast-based probe approach, we address the inconsistency and locate two critical regions for effective knowledge editing.
$\left ( 2 \right ) $ We propose JEEP, a knowledge editing method that jointly edits both the low and high layers to modify the stages of information enrichment and probability promotion, offering a more holistic and effective approach to knowledge editing.
$\left ( 3 \right ) $ We conduct extensive comparative experiments across various numbers of edits, model architectures, and datasets. These experiments not only demonstrate the effectiveness of our JEEP method but also highlight the practical values of our probe findings in designing knowledge editing methods.

\section{Related Work}

\subsection{Probe into Knowledge Recall Process}
Probe approaches into knowledge recall process try to elucidate a model's internal processing of the output answers \cite{mechanism,gurnee2023finding,bills2023language}, where our focus is on critical regions for knowledge editing \cite{katz2023interpreting}. It is widely believed that the Multi-Layer Perceptron (MLP) module serves as a primary repository of knowledge \cite{geva2020transformer,dai2021knowledge,kobayashi2023feed}, and numerous studies analyze its behavior as key-value memories \cite{geva2022transformer,dar2022analyzing}. By restoring corrupted hidden states, \cite{meng2022locating} reveals that early-to-middle MLP layers enrich subject-related factual information in representations of subject's last token, thereby pinpointing a more specific knowledge storage location. Regarding the Multi-Head-Self-Attention (MHSA) module, \cite{dar2022analyzing} shows that MHSA parameters also encapsulate knowledge. Further, \cite{geva2023dissecting} blocks the attention computations to trace detailed information flows of final predictions, underscoring MHSA's role in extracting answers to the prediction position. 
These findings outline the critical stages of the knowledge recall process. However, the used ablation-based probes only and could only reveal critical stages for the original answers, failing to provide complete critical regions for knowledge editing, which aims to edit for the target answers. By observing and contrasting the original and target answers' information changes across layers, we endeavor to resolve the inconsistency and provide a more comprehensive understanding of the knowledge recall process for effective editing.

\subsection{Knowledge Editing}
Knowledge editing methods can be categorized into two lines based on whether they modify model parameters. Weight-preserved methods incorporate additional structures to handle incoming editing demands, such as new neurons \cite{huang2023transformer}, scope classifiers with counterfactual models \cite{mitchell2022memory}, adapters \cite{hartvigsen2024aging}, LoRA modules \cite{yu2024melo}, and in-context learning examples \cite{zheng2023can}. But they are not scalable, as they become more costly and less effective with an increasing number of edits. In contrast, weight-modified methods directly update the model's weights, enabling large-scale simultaneous edits. The most straightforward approach, constrained fine-tuning, suffers from severe overfitting issues \cite{zhu2020modifying}. Meta-learning-based methods train hyper-networks to generate parameter updates that satisfy the generalization and locality of editing, yet they struggle to maintain performance at larger editing scales \cite{de2021editing,mitchell2021fast,tan2023massive}. Locate-then-edit methods \cite{meng2022mass, li2024pmet}  leverage findings from probe approaches into knowledge recall process \cite{meng2022locating,geva2023dissecting}, locating low-layer MLPs and editing them as key-value memories. While these methods could handle extensive edits in one operation, the inconsistency between the probes and editing targets limits their effectiveness. Typically, they modify only partial critical regions and fail to align the final predictions with the updated information  effectively. To address these challenges, we propose to edit both the low and high layers to modify the information enrichment and probability promotion stages based on our probe revealings. This approach leads to a more thorough modification of the knowledge recall process and  improved editing performance.

\section{Preliminaries}
\subsection{Language Modeling}
In a decoder-only language model $\mathcal{F_{\theta}}$ \cite{brown2020language}, the input sequence $[\mathrm {x}_1, \mathrm {x}_2,...,\mathrm {x}_{E}] $ goes through \textit{D}-layer computations, and the last token representation in the final layer $h_E^D$ would be mapped to the vocab distribution through the language model (LM) head $W_{lm}$, to decode the probabilities of the next token $\mathrm {x}_{E+1}$:
\begin{equation}
\begin{aligned}
\mathcal{F_{\theta}}\left (\left [ \mathrm {x}_{1},\mathrm {x}_{2}\ldots \mathrm {x}_{E}  \right ]\right)\triangleq \mathbb{P}_{E}^{\textit{D}}=softmax(W_{lm}(h_{E}^{\textit{D}})).
\label{lm_final}
\end{aligned}
\end{equation}
So the representation $h_{E}^{\textit{D}}$ encodes information of the next token. This information is accumulated during \textit{D}-layer residual connections:
\begin{equation}
\begin{aligned}
h_{E}^{D}= h_{E}^{0}+&\sum_{l=1}^{\textit{D} } (a_{E}^{l}+m_{E}^{l}),\\
where \ &a_{E}^{l}=  W_{O^{\mathrm{MHSA}}}^{l}\mathrm{MHSA}(\gamma(h_{1}^{l-1},h_{2}^{l-1},...,h_{E}^{l-1} ))\\
&m_{E}^{l} = W_{O^{\mathrm{MLP}}}^{l}\sigma (W_{IN^{\mathrm{MLP}}}^{l} \gamma (h_{E}^{l-1})),
\label{residual}
\end{aligned}
\end{equation}
and $h^0$ is the embedding and $\gamma$ denotes layer normalization.

\subsection{Knowledge Editing}\label{KE_def}
LLM $\mathcal{F}$ has encoded abundant knowledge in its parameters:
\begin{equation}
\begin{aligned}
\textit{K}_{\mathcal{F}} = \left \{(x_i,y_i)_{i=1}^{N},  \mathcal{F}(x_i)=y_i\right \},
\label{lm_layer}
\end{aligned}
\end{equation}
where $\textit{K}_{\mathcal{F}}$ is the \textbf{original knowledge} in the model and $(x_i,y_i)$ denotes a input prompt and answer pair. Given a knowledge pair $\left ( x,y \right ) $, such as (World Cup's winner is, France), there are two responding sets: Equivalent Set $\textit{E}\left ( x,y \right ) $, containing all semantically equivalent knowledge pairs to $\left ( x,y \right ) $, an example would be (Who is the World Cup's winner?, France); Unrelated Set $\textit{U}\left ( x,y \right ) $, containing all unrelated knowledge pairs to $\left ( x,y \right ) $, an unrelated pair would be (Which country does Paris belong to?, France).

For $m$ pieces of \textbf{target knowledge} pairs $(x_{i},y_{i}')_{i=1}^{m}$ to be edited, knowledge editing aims to change model's predictions to the target answers on these inputs (Efficacy), while generalizing the model to their equivalent pair collection $ \bigcup_{i=1}^{m} \textit{E}(x_i,y_i') $ (Generalization) , and maintaining the predictions on the unrelated pair collection $\bigcap_{i=1}^{m} \textit{U}(x_i,y_i') $ (Locality). Therefore, the edited model $\tilde{\mathcal{F}}$ should simultaneously satisfy the following three goals:
\begin{equation}
\begin{aligned}
\tilde{\mathcal{F}}(x)&=y \  \wedge\  \tilde{\mathcal{F}}(x^e)=y^e  \ \wedge \  \tilde{\mathcal{F}}(x^u)=y^u, \\
(x,y)&\in \left \{ (x_{i},y_{i}')_{i=1}^{m} \right \} ,(x^e,y^e)\in \bigcup_{i=1}^{m} \textit{E}, (x^u,y^u)\in \bigcap_{i=1}^{m} \textit{U}.
\label{lm_layer}
\end{aligned}
\end{equation}
Detailed measuring metrics are in Appendix \ref{APP: dataset}.

\section{Method}\label{method}
Exploring the knowledge recall process aids in designing effective knowledge editing methods \cite{meng2022mass,li2024pmet}. But existing probes could only locate prominent regions for original answers, which is inconsistent to the knowledge editing for target answers. To mitigate this, we propose a contrast-based probe approach and identify two critical recall stages for editing (\S \ref{probe}).
Based on our discoveries, we develop JEEP to more thoroughly modify the knowledge recall process to alter model predictions, while addressing the issues associated with joint updates (\S \ref{edit_both}).

\subsection{Unveiling Critical Editing Stages}\label{probe}
Our probe contrasts the detailed information flow of the original and target answers across different representations. Guided by the probe, we identify two critical editing stages: Information Enrichment in low layers and Probability Promotion in high layers.

\subsubsection{The Contrast-Based Probe Approach}  
\begin{figure}[t]
\centering
\includegraphics[scale = 0.252]{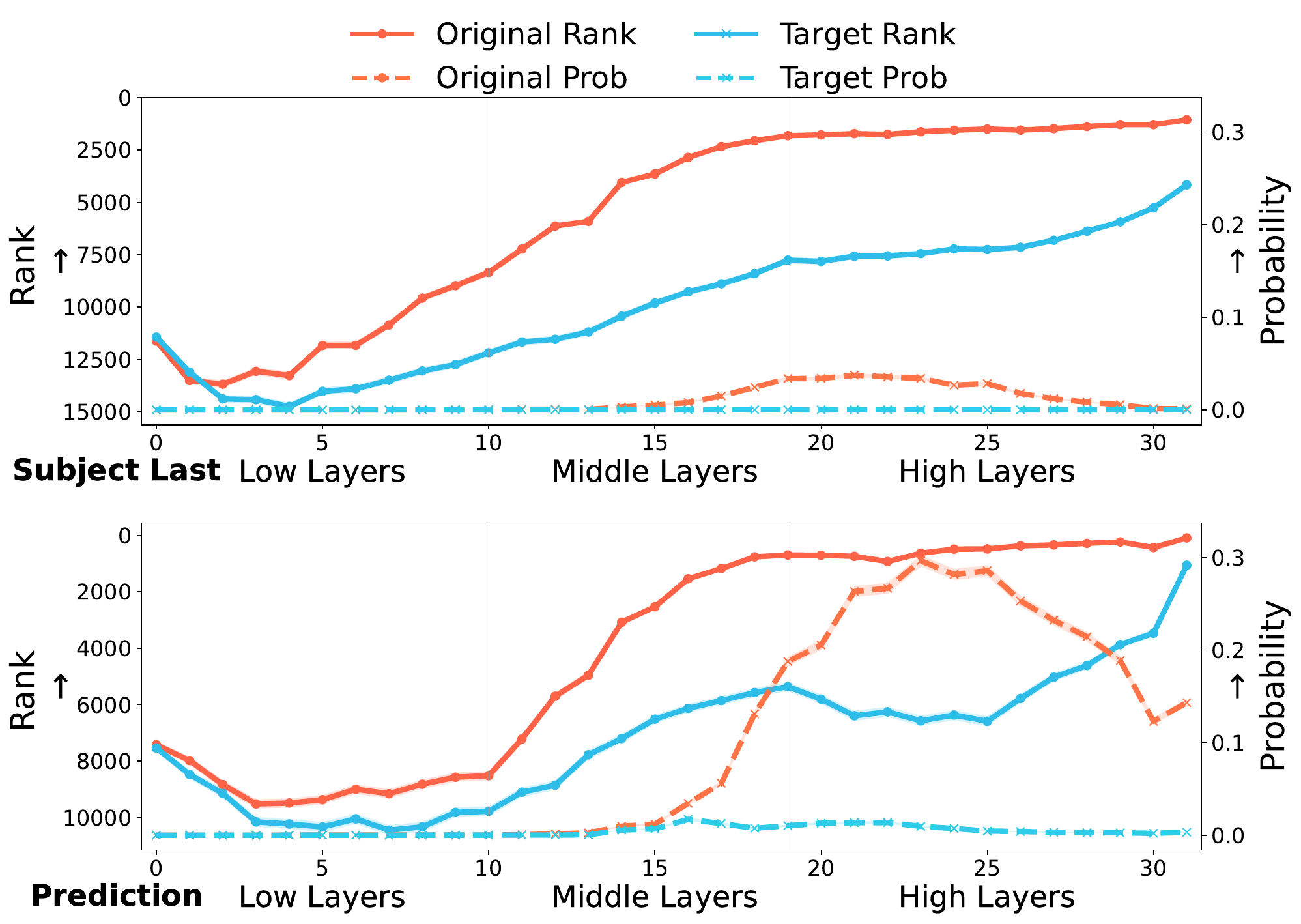}
\caption{Original and Target answers' information change in low, middle and high layers, indicated by rank and probability (Prob). The top and bottom graphs are for Subject Last and Prediction positions respectively.} 
\label{pic_probe}
\end{figure}

Since previous ablation-based probes fail to analyze the target answers, we directly observe and contrast the differences in the information flow between the original and target answers. Extended from Eq. \ref{residual}, representation at any position $i$ and any layer $l$ is refined by residual connections. So the LM head's mapping could be applied to any representation $h_{i}^{l}$ \cite{logitlens}:
\begin{equation}
\begin{aligned}
\mathbb{P}_{i}^{l} = softmax(W_{lm}(h_{i}^{l})).
\label{lm_layer}
\end{aligned}
\end{equation}
Recording the probability $\mathbb{P}(t)$ of the answer's first token $t$ within the distribution $\mathbb{P}$, we can observe the complete information flow of answers inside the model. But probability only reflects absolute information and are often very low for the target answers. Therefore, we further calculate the rank of the target token $t$ in the distribution:
\begin{equation}
\begin{aligned}
\text{rank}(t |\mathbb{P}) = \sum_{t' \neq t} \mathbf{1}(\mathbb{P}(t') > \mathbb{P}(t)) .
\label{lm_layer}
\end{aligned}
\end{equation}
Note that the subsequent ``increase" for rank indicates the rank value is approaching 0, and the representation contains more information about the token. In Fig. \ref{pic_probe}, we input 10,000 prompts $x$ into LLaMA-7B model, recording average probability and rank of the first token in the original answers $y$ and target answers $y'$ for editing (see Appendix \ref{APP: prob} for details). To differentiate information flow in different positions, we decompose the prompt-answer knowledge pair  $\left ( x,y \right ) $ into the form of a triplet $\textless subject, relation, object\textgreater$, the most common format for representing knowledge \cite{petroni2019language}. For example, (World Cup's winner is, Argentina) can be decomposed into \textless World Cup, 's winner is, Argentina\textgreater. And we focus on representations at subject's last and prediction positions, since the former encapsulates all the subject information \cite{meng2022locating}, and the later is where the prediction is made.

By contrasting the differences in information flow between the original and target answers, we identify critical stages with significant model behavior differences, which are suitable for knowledge editing. As depicted in Fig. \ref{pic_probe}, these stages align with the low, middle, and high layers of the model.
In the low layers, at the subject's last position, the rank of original answers increases, marking an \textbf{Information Enrichment} stage, mainly driven by the MLP module \cite{meng2022locating}. Target answers, however, do not experience similar enrichment, as their rank only increases marginally, highlighting this stage's importance for effective knowledge editing.
In the middle layers, information for both original and target answers at the prediction position starts to increase, evidenced by a sharp rise in rank, primarily due to the MHSA module \cite{geva2023dissecting}. This indicates that the MHSA consistently extracts all potential answers fitting the relation, including the target answers, suggesting a stable extraction pattern. Therefore, this stage requires no modification during editing, aligning with previous editing methods \cite{meng2022mass,li2024pmet}.
Finally, in the early high layers at the prediction position, original answers undergo \textbf{Probability Promotion}. Although this probability increase might seem minor due to the slight rank rise of original answers at this stage, the rank of target answers, lacking probability promotion, decreases instead of increasing. This contrast highlights the crucial role of Probability Promotion in elevating original answers to become the top-ranked output, indicating that this stage should be modified for effective knowledge editing.

Besides, we note a decline in the probability of original answers at the prediction position in the final layers, and even the rank decreases in the penultimate layer. To explore it, we expand our analysis to include additional word sets, examining their information changes. We discover that the uppermost layers function as distribution normalizers, increasing information entropy at the prediction position and aligning outputs with realistic data distribution. So the early high layers are more conducive for editing. We leave the detailed analysis in Appendix \ref{APP: prob}.

\subsection{Joint Knowledge Editing}\label{edit_both}
Based on the observations above, we develop the JEEP method, which edits both the low and high layers, specifically targeting the MLP modules therein, to simultaneously modify the stages of information enrichment and probability promotion. 
Initially, we outline the comprehensive process of the joint editing. Subsequently, we present two challenges associated with dual updates, mutual interference and increased forgetting, and our tailored designs to solve them.

\begin{figure}[t]
\centering
\includegraphics[scale = 0.3]{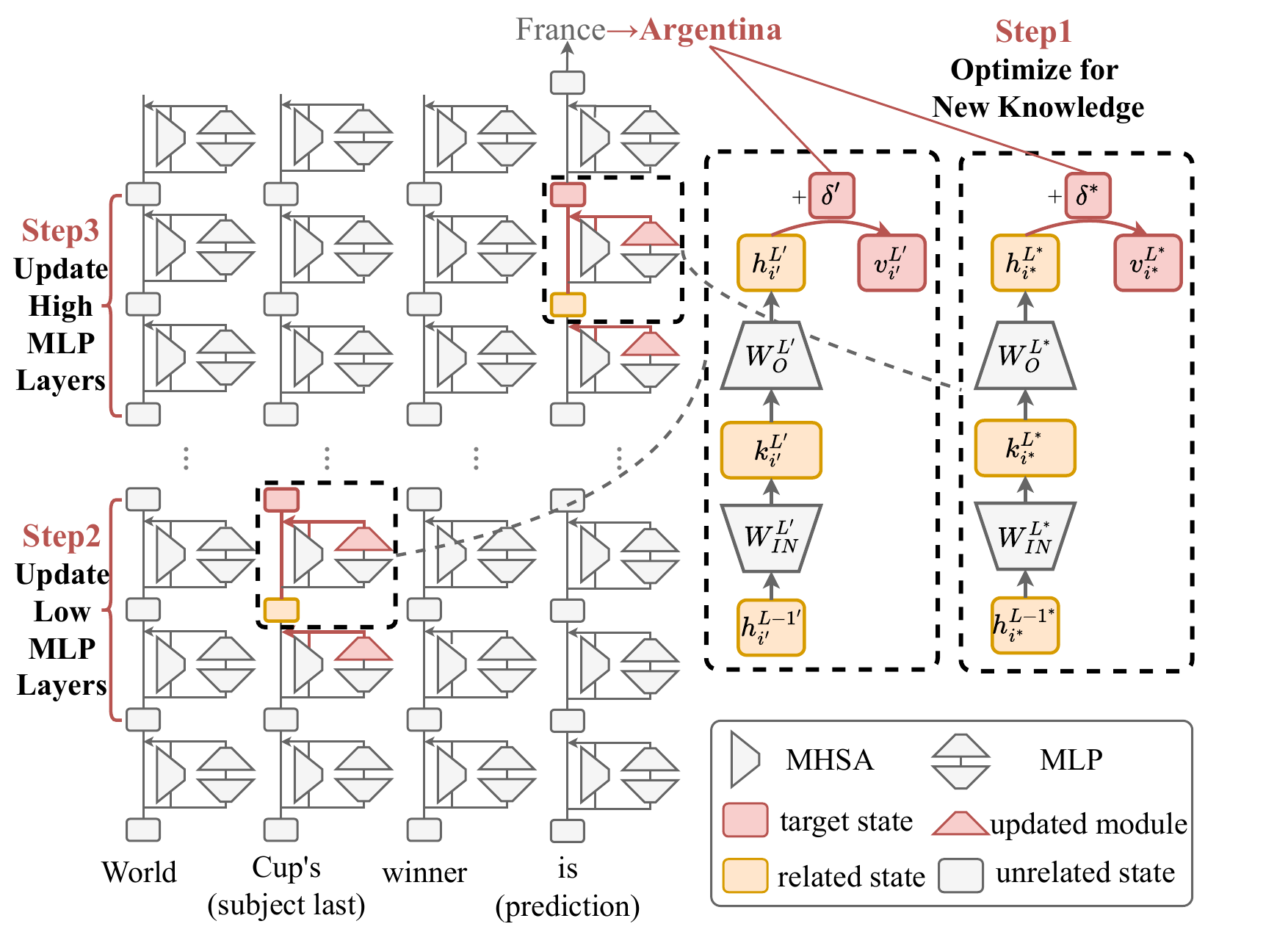}
\caption{Procedure of JEEP method. Firstly, it computes $\delta'$ and $\delta^{*}$ for low and high layers simultaneously, optimizing for injecting new knowledge. Secondly, it uses the residual errors of $v_{i'}^{L'}$ to update the low MLP layers. Finally, it uses the residual errors of $v_{i^{*}}^{L^{*}}$ to update the high MLP layers.} 
\label{method}
\end{figure}
\subsubsection{Joint Editing Procedure} To revise the knowledge recall process more completely, we edit the MLP modules in both low layers $\left [l',L' \right ] $ and high layers $\left [l^*,L^* \right ] $. We analyze the MLP module as a key-value memory, where $W_{IN}$ encodes input into key vectors and $W_{O}$ maps keys to output values containing knowledge. Suppose the model memorizes $n$ pieces of original knowledge $(x_{i},y_{i})_{i=1}^{n}$, we have $n$ mappings encoded in $W_{O}$:
\begin{equation}
\begin{aligned}
W_{O}K_0=V_0, \ 
\label{prob}
\end{aligned}
\begin{aligned}
K_0 &\triangleq [k_1\mid k_2 \dots \mid k_n ]\\
V_0 &\triangleq [v_1\mid v_2 \dots \mid v_n ].
\end{aligned}
\end{equation}
To edit $m$ knowledge pairs, we update the model in three steps as shown in Fig. \ref{method}. Firstly, for every target knowledge pair $\left ( x,y' \right ) $, we calculate the corresponding value vector $v_{i}^{L}$ ($i \in \left \{ i', i^* \right \}, L \in \left \{ L',L^* \right \}$ ) to replace the current hidden state $h_i^L$, at subject last position $i'$ and prediction position $i^*$  in last critical layer $L'$ and $L^*$. We optimize the residual vector $\delta$ ($\delta \in \left \{ \delta',\delta^* \right \}$  ) by gradient descend to alter model's prediction on $x$ to $y'$ (Step1):
\begin{equation}
\begin{aligned}
v_{i}^{L} =  h_{i}^{L} + \mathop{argmin}_{\delta}\frac{1}{P}\sum_{j=1}^{P}\mathcal{L}_{\mathcal{F}(h_{i}^{L}+=\delta)}(y'\mid p_{j}+x).
\label{value}
\end{aligned}
\end{equation}
Different prefixes $p_{j}$ are used to bolster generalization. We formulate the editing to adding $m$ new mappings while preserving the existing $n$ ones with minor change $\Delta$ to $W_O$:
\begin{equation}
\begin{aligned}
(W_O+\Delta) [K_0 K_1]=[V_0 V_1], \ 
\label{prob}
\end{aligned}
\begin{aligned}
K_1 &\triangleq [k_{n+1}\mid  \dots \mid k_{n+m} ]\\
V_1 &\triangleq [v_{n+1}\mid \dots \mid v_{n+m} ] .
\end{aligned}
\end{equation}
By derivation in \cite{meng2022mass}, we could obtain:
\begin{equation}
\begin{aligned}
\Delta=RK_{1}^{T}(C_{0}+K_{1}K_{1}^{T})^{-1},
\label{memit}
\end{aligned}
\end{equation}
where $R\triangleq V_{1}-W_{O}K_{1}$, denoting the residual errors when assessing new knowledge mappings on the original weights. And $C_{0}\triangleq K_{0}K_{0}^{T}$ denotes the uncentered covariance of the original key vectors. Lacking access to the input prompts for the original knowledge $K_0$, we instead sample prompts from Wikipedia and estimate $C_0$ based on their key vectors to $W_{O}$: $C_{0}\triangleq \lambda \cdot \mathbb{E}_k  \left [kk^{T}  \right ] $. 
Utilizing Eq. \ref{memit}, we initiate updates in the lower region (Step2) before advancing to the upper region (Step3). Within each region, updates proceed sequentially from the lower to the higher layers. Specifically, for the current updating layer $l$, the key vector $k_{i}^{l}$ is calculated by averaging the input keys of editing prompts with different prefixes: $k = \frac{1}{P}\sum_{j=1}^{P}k(p_{j}+x)$. To distribute the updates evenly across all targeted layers, we spread the current residual error by the number of layers pending update: $r = \frac{v_{i}^{L}-h_{i}^{L}}{L-l+1}$. Horizontally stacking different $k$ and $r$ corresponding to new knowledge pairs, we obtain $K_1$ and $R$ and compute the matrix update by Eq. \ref{memit}.

\subsubsection{Synergistic Optimization}
The first challenge in joint editing is mutual interference between updates across different regions. To harmonize the dual updates, we compute the value vectors in Eq. \ref{value} simultaneously, ensuring they are optimized in a synergistic manner. The optimization objective is defined as follows:
\begin{equation}
\begin{aligned}
\mathcal{L_{}}(\delta',\delta^{*}) &= \mathrm {-log} {\mathbb{P}_{n}^{D}}_{\mathcal{F}(h_{i'}^{L'}+=\delta',h_{i^{*}}^{L^{*}}+=\delta^{*})}\left [ y'\mid x\right ] \\ 
&+\beta'*\left \| \delta' \right \| +\beta^{*}*\left \| \delta^{*} \right \|\\
&+\alpha' *D_{kl}({\mathbb{P}_{i}^{D}}_{\mathcal{F}(h_{i'}^{L'}+=\delta')}\left [ x\right ] \mid\mid {\mathbb{P}_{i}^{D}}\left [ x\right ]),
\label{loss_both}
\end{aligned}
\end{equation}
where $\beta'$ and $\beta^*$ are coefficients for the weight decay loss of $\delta'$ and $\delta^*$, respectively. And $\alpha'$ adjusts the KL-divergence loss for output at the subject's last token. In addition to the language model loss, which is used to inject new knowledge into the model, we further incorporate KL-divergence and weight decay terms to mitigate forgetting, ensuring minimal disruption to other knowledge.

\subsubsection{Adaptive Updates}
The second challenge encountered by a joint editor is increased forgetting due to more parameter updates within the model. To adjust the extent of the updates, we compress the knowledge encoded in the modifications, by clamping the L2 norms of $\delta$ and $\delta'$ based on their respective original hidden state norms: 
\begin{equation}
\begin{aligned}
\left \| \delta' \right \| <= \gamma' * \left \| h_{i'}^{L'} \right \|,
\left \| \delta^{*} \right \| <= \gamma^{*} * \left \| h_{i^{*}}^{L^{*}} \right \|,
\label{loss_both_ada}
\end{aligned}
\end{equation}
where $\gamma'$ and  $\gamma^*$ are coefficients to control the upper bounds for $\delta'$ and $\delta^*$, respectively.
We find that modifications in the upper layers of model have a greater impact on the final outputs, suggesting that the required change in probability promotion to alter predictions is less than that for information enrichment (see Appendix \ref{APP: analysis}). Thus, more updates should be allocated to the low layers. Coupled with the fact that the lower state norm is smaller, we set: $\gamma' > \gamma^{*}$. Furthermore, since the knowledge encoded in low layers is denser, we spread the residual errors as follows: $r = \frac{v_{i'}^{L'}-h_{i'}^{L'}}{\sqrt{L'-l'+1}}$ \cite{li2024pmet}, enhancing the updates. By adaptively adjusting the update magnitudes in these two regions, we effectively alter model's predictions with minimal parameter changes.

\begin{table*}[t]
\centering
\begin{tabular}{ccccc}
\hline
            & Input Prompt                                                                                                              & MEMIT & PMET & JEEP \\ \hline
Edited& \textit{Where was \textbf{Henry S. LeBlanc} from?}    & Canada   & Canada  & Canada  \\  

Equivalent  & \textit{Where did \textbf{Henry S. LeBlanc} come from?} & \textcolor{red}{of course}     & Canada  & Canada  \\
Unrelated   & \textit{Where does \textbf{creatine} come from in the body?}                                                                                  & liver   & \textcolor{red}{Where is creatine from?} 
    & liver \\ \hline
\end{tabular}
\caption{A case of three editors' predictions on the prompts for evaluating efficacy, generalization and locality. Input Prompt consists of the subject, identified in bold, and the relation. Answers in red are inaccurate, either meaningless or repetitive. }
\label{case}
\end{table*}

\begin{table}[t]
\centering
\begin{tabular}{lcccc}
\hline
\textbf{Method}  &\textbf{Score}$\ \uparrow$ & \textbf{ES}$\ \uparrow$ & \textbf{GS}$\ \uparrow$ & \textbf{LS}$\ \uparrow$ \\ \hline
GPT-J    &26.4 & 26.4 (0.6)  & 25.8 (0.5)       & 27.0 (0.5)    \\ \cdashline{1-5}

FT-WD      & 42.1 & 69.6 (0.6)      & 64.8 (0.6)  & 24.1 (0.5) \\
MEND      &20.0   & 19.4 (0.5)  & 18.6 (0.5)       & 22.4 (0.5)      \\
ROME      &2.6   & 21.0 (0.7)  & 19.6 (0.7)       & 0.9 (0.1)      \\
MALMEN      &47.1   & 98.3 (0.3) & 90.1 (0.2)      & 23.6 (0.4)    \\
MEMIT     &50.7  & 96.7 (0.3)      & 89.7 (0.5)      & 26.6 (0.5)    \\
PMET      &51.0   & 96.9 (0.3) & 90.6 (0.2)      & 26.7 (0.2)    \\
JEEP      &\textbf{51.5}   & \textbf{98.4} (0.2) & \textbf{91.5} (0.3)      & \textbf{26.9} (0.2)    \\ \hline
LLaMA    &44.4 & 43.6 (0.3)  & 42.6 (0.5)       & 49.2 (0.6)    \\ \cdashline{1-5}

MALMEN      &66.7   & 90.1 (0.4) & 84.6 (0.3)      & 45.3 (0.5)    \\
MEMIT     &70.4  & 95.1 (0.2)      & 89.5 (0.4)      & 47.8 (0.5)    \\
PMET      &69.2   & 91.3 (0.3) & 86.4 (0.5)      & 48.0 (0.3)    \\
JEEP      &\textbf{72.3}   & \textbf{96.0} (0.2) & \textbf{93.2} (0.4)      & \textbf{49.1} (0.2)    \\ \hline 
\end{tabular}
\caption{Results on zsRE for 10,000 edits on GPT-J (6B) and LLaMA (7B) models.}
\label{res_zsre}
\end{table}

\section{Experiment}
\subsection{Experimental Setup}
We conduct experiments on two models, GPT-J (6B) \cite{wang2021gpt} and LLaMA (7B) \cite{touvron2023llama}, which feature parallel and sequential MHSA and MLP modules, respectively. 
Based on the three objectives of knowledge editing, we evaluate three corresponding key metrics: \textbf{Efficacy Success} (ES), \textbf{Generalization Success} (GS) and \textbf{Locality Success} (LS), then compute their harmonic mean \textbf{Score} as in \cite{meng2022mass} (see Appendix \ref{APP: dataset} for detailed descriptions).
For baselines, we first consider \textbf{FT-WD}, fine-tuning with weight decay to prevent forgetting \cite{zhu2020modifying}. Next, for meta-learning-based methods, we include \textbf{MEND} and its improved version \textbf{MALMEN}, which further supports multiple facts' editing at once \cite{mitchell2021fast,tan2023massive}. Finally, we compare with locate-then-edit methods: \textbf{ROME}, \textbf{MEMIT} and \textbf{PMET} \cite{meng2022locating,meng2022mass,li2024pmet}. Since \textbf{ROME} originally targets one fact at a time, we adapt it to a sequential version to accommodate massive edits \cite{meng2022mass}. We evaluate editors' ability to add factual knowledge (\S \ref{add_factual}), and further evaluate the capability to add counterfactual knowledge (\S \ref{add_counterfactual}), which is a more challenging scenario. Implementation details are in Appendix \ref{APP: imple}.

\subsection{Adding Factual Knowledge}\label{add_factual}
\subsubsection{Editing 10k Samples}
The primary goal of \textit{Knowledge Edit} is to correct inaccuracies in the model's original knowledge. Initially, we evaluate the editing methods' abilities to add \textit{factual} knowledge. We extract 10,000 real-world factual pairs $\left ( x,y' \right ) $ from zsRE \cite{levy2017zero}, a question-answering dataset. As the editing targets are correct answers, all three metrics compute the proportion of tokens in $y'$ that have the highest probability given the input prompt $x$ (see Appendix \ref{APP: dataset}). As shown in Table \ref{res_zsre}, not all methods can handle such a large volume of edits, highlighting the complexity of editing 10,000 samples concurrently. Some baseline methods, such as MEND and ROME, perform even worse than simple fine-tuning. The JEEP method shows superior performance across all three metrics on both models and further improvements on the advanced, knowledge-rich LLaMA model, validating its editing designs.
And compared to the low-layers-focused methods MEMIT and PMET, JEEP's improvements emphasize the benefits of a more comprehensive understanding and modification of the knowledge recall process in editing methods.
Moreover, the 95\% confidence intervals of results confirm that JEEP achieves statistically significant improvements over the baselines, with consistently narrow intervals, highlighting its robustness.

\subsubsection{Case Study} 
In Table \ref{case}, we showcase the predictions of the LLaMA model, after being edited by different editors, on three metrics for one of the 10,000 editing samples. All editors make accurate predictions on the edited prompt, indicating that achieving basic editing efficacy is feasible. However, only JEEP performs properly on both equivalent and unrelated prompts, suggesting incomplete modifications to the knowledge recall process result in either insufficient or overfitting editing.

\begin{table}[t]
\centering
\begin{tabular}{lcccc}
\hline
\textbf{Method}  &\textbf{Score}$\ \uparrow$ & \textbf{ES}$\ \uparrow$ & \textbf{GS}$\ \uparrow$ & \textbf{LS}$\ \uparrow$ \\ \hline
GPT-J     & 22.4 & 15.2 (0.7) & 17.7 (0.6)      & 83.5 (0.5)   \\ \cdashline{1-5}
FT-WD      & 67.6 & 99.4 (0.1) & 77.0 (0.7)      & 46.9 (0.6)    \\
MEND      & 23.1  & 15.7 (0.7) & 18.5 (0.7)       & \textbf{83.0} (0.5)    \\
ROME      & 50.3  & 50.2 (1.0) & 50.4 (0.8)       & 50.2 (0.6)    \\
MALMEN      & 63.4  & 98.2 (0.8) & 46.0 (0.8)       & 65.1 (0.6)    \\
MEMIT     & 85.8 & 98.9 (0.2) & 88.6 (0.5)      & 73.7 (0.5)    \\
PMET     & 86.2 & 99.5 (0.1) & \textbf{92.8} (0.4)      & 71.4 (0.5)    \\
JEEP         & \textbf{86.5}  & \textbf{99.8} (0.1) & 90.9 (0.5)      & 73.1 (0.5)    \\ \hline
LLaMA     & 19.7 & 12.9 (0.6) & 16.0 (0.7)      & 83.8 (0.4)    \\ \cdashline{1-5}
MALMEN      & 66.1  & 96.0 (0.3) & 55.3 (0.8)       & 59.3 (0.8)    \\
MEMIT     & 84.1 & 98.4 (0.2) & 80.6 (0.6)      & \textbf{76.3} (0.5)    \\
PMET         & 84.8 & 98.9 (0.2) & \textbf{86.2} (0.5)      & 73.2 (0.6)    \\ 
JEEP         & \textbf{85.9} & \textbf{99.8} (0.1) & 85.4 (0.5)      & 75.7 (0.5)    \\ \hline 
\end{tabular}
\caption{Results on Multi-COUNTERFACT for 10,000 edits on GPT-J (6B) and LLaMA (7B) models. }
\label{res_mcf}
\end{table}
\begin{figure*}[t]
\centering
\includegraphics[scale = 0.42]{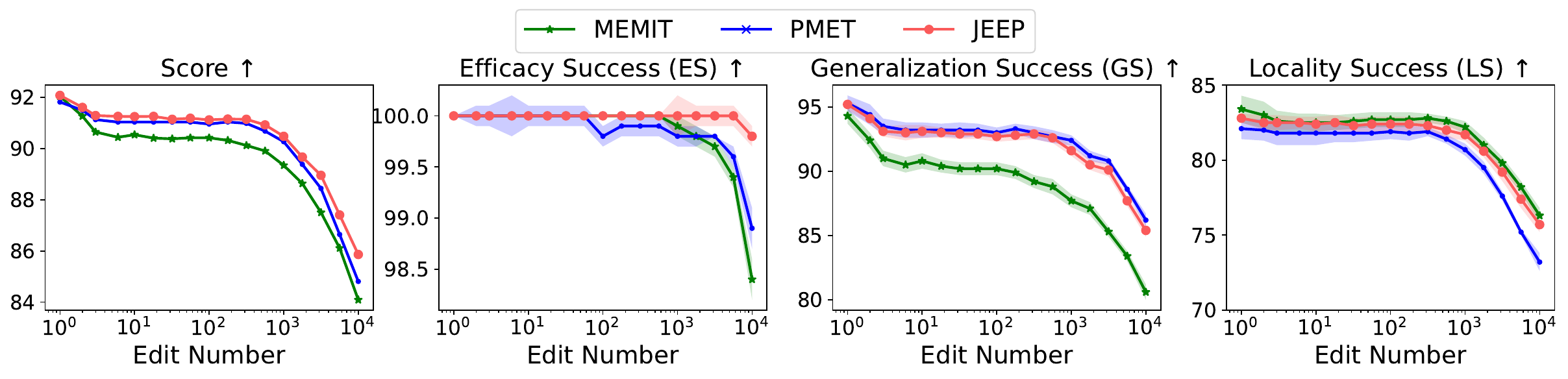}
\caption{Scaling curves plot performance change against different editing numbers (log-scale) on LLaMA (7B). 95\% confidence intervals are shown as areas. } 
\label{edit_num}
\end{figure*}

\subsection{Adding Counter-Factual Knowledge}\label{add_counterfactual}
\subsubsection{Editing 10k Samples}
We next shift to a more challenging scenario: evaluating methods' capabilities in injecting \textit{counterfactual} information. We edit 10,000 samples from the Multi-COUNTERFACT dataset \cite{meng2022mass}, where the editing updates original correct answers to target incorrect answers. 
The evaluation for successful editing is more lenient, and the metrics calculate the proportion of cases where the generation probability of the target incorrect answer exceeds that of the original correct answer (Appendix \ref{APP: dataset}). But as for locality, a higher probability of the original correct answer is considered successful.
Given the difficulty for adding this artificially constructed  erroneous knowledge, it's challenging for editing methods to excel in the efficacy, generalization, and locality simultaneously. 
As illustrated in Table \ref{res_mcf}, no method consistently exceeds the others across all metrics. In this unnatural setting, our JEEP method achieves the best efficacy and near-best generalization and locality, leading to optimal overall performance. This indicates that more comprehensive modifications of critical knowledge recall stages are also beneficial for adding counterfactual knowledge, validating the effectiveness of our probe findings and editing method.

\subsubsection{Editing Scaling}
To evaluate the scalability of our method, we compare the performance of methods capable of handling 10,000 edits as the number of edits $m$ increases\footnote{$m$ is sampled from a log-scale curve: $m_{i}=exp(ln(10,000)*\frac {i}{16})$, for non-negative integers $i$.}. As shown in Fig. \ref{edit_num}, JEEP consistently delivers better overall results, with the improvements becoming more pronounced as the number of edits grows. 
Specifically, JEEP achieves 100\% accuracy in all cases except for 10,000 edits and remains near the best in generalization and locality, demonstrating its robust performance across varying edit numbers.

\begin{table}[t]
\centering
\begin{tabular}{lcccc}
\hline
\textbf{Variant}  &\textbf{Score}$\ \uparrow$ & \textbf{ES}$\ \uparrow$ & \textbf{GS}$\ \uparrow$ & \textbf{LS}$\ \uparrow$ \\ \hline
w/o $\delta' \& $ Step2 & 66.3 & 89.8 & 56.6      & 60.7 \\
w/o $\delta^{*} \& $ Step3 & 84.1 & 98.4& 80.6  & \textbf{76.3} \\ 
w/o Step2 & 40.0 & 36.5 & 24.1      & 81.5 \\ 
w/o Step3 & 81.9 & 95.4 & 77.5      & 75.4 \\ \hline

Separate Optimization &83.3 & 98.2 & 83.3      & 72.5 \\
Even Spread in Step2 &85.4 & 98.8 & 84.0      & \textbf{76.3} \\
w/ $\mathrm {MHSA}$ in Step3 & 85.2 & 99.0 & \textbf{86.6}      & 73.8 \\
JEEP         & \textbf{85.9} & \textbf{99.8} & 85.4      & 75.7 \\
\hline
\end{tabular}
\caption{Effects of ablating different designs of JEEP. The top half focuses on single-region editing. The bottom half ablates different designs within JEEP.}
\label{abla}
\end{table}

\subsection{Ablation Study} 
To elucidate the contributions of updates from different regions and various designs in JEEP, Table \ref{abla} presents the results of different ablation variants, editing 10,000 samples from Multi-COUNTERFACT on LLaMA.

$1 )$ Initially, to assess JEEP's design of dual-region modifications, we examine single-region editing. In \textbf{w/o $\delta' \& $} \textbf{Step2} and \textbf{w/o}  $\mathbf{\delta^{*}\&} $ \textbf{Step3},  only one residual vector in Eq. \ref{value}, either $\delta^*$ or $\delta'$, is computed and used to update the high or low layers respectively. They both underperform compared to JEEP, underscoring the advantage of updating both regions. Notably, editing the lower layers along performs better, suggesting that without the information enrichment from the lower MLP, modifying probability promotion alone is less effective. 
Additionally, \textbf{w/o Step2} or \textbf{w/o Step3} keeps the simultaneous computation of $\delta'$ and $\delta^*$, but subsequently updates only one region. This approach considers the interplay between updates but limits the actual modification to one regions. As with previous findings, single-region editing yields inferior results compared to JEEP, with lower-layer editing still shows better outcomes. However, compared to computing only one $\delta$, the dual computations of $\delta$ performs worse, suggesting a more holistic consideration of the critical recall stages without corresponding updates may instead harm the editing effect. JEEP improves this by adaptively updating both regions, achieving a more balanced distribution of information changes and superior performance.

$2 )$ Moreover, we ablate the designs in JEEP. \textbf{Separate Optimization} deviates from JEEP's synergistic design in Eq. \ref{loss_both}, by independently computing $\delta'$ and updating the low layers before computing  $\delta^*$ and updating the high layers. Its performance is not only inferior to JEEP but also worse than single-region editing, indicating that isolating the optimization of two regions leads to mutual interference between them. In addition, \textbf{Even Spread in \textbf{Step2}} replaces the excessive updates to the low layers by spreading the residual errors evenly: $r = \frac{v_{i'}^{L'}-h_{i'}^{L'}}{L'-l'+1}$. It performs worse than JEEP, achieving better locality at the cost of efficacy and generalization, but still outperforms updating one region. 
Furthermore, considering the computational similarities between $W_O$ in MLP and  MHSA, \textbf{ w/ $\mathrm {\textbf{MHSA}}$ in Step3} instead edits MHSA in high layers. Although this adjustment does not outperform JEEP, it shows improved performance over single-region editing, indicating that both MLP and MHSA could promote answer's probability in the upper layers.

These ablation studies not only validate the insights from our probe experiments but also demonstrate the effectiveness of the design choices in JEEP.

\section{Conclusion}
In this study, we discover the inconsistency between existing probe approaches for the original answers and the knowledge editing goal for the target answers. 
To address the inconsistency, we propose a contrast-based probe approach, identifying two critical knowledge recall stages for editing: Information Enrichment in low layers and Probability Promotion in high layers.
Based on these insights, we develop  JEEP, a joint editing method targeting both low and high layers to modify the information enrichment and probability promotion stages simultaneously. Through synergistic optimization and adaptive updates, JEEP addresses the challenges of mutual interference and increased forgetting  associated with updating different model regions, consistently achieving superior performance across various edit scales, models, and datasets.

\section*{Acknowledgements}
The work was supported by the Outstanding Innovative Talents Cultivation Funded Programs 2023 of Renmin University of China and the National Natural Science Foundation of China under Grant No. 62072458.

\bibliography{aaai25}

\clearpage 
\section*{Appendix}
\appendix

\section{Probe on Knowledge Recall Process}\label{APP: prob}
In \S 4.1, we investigate the fine-grained information changes of the original and target answers during the knowledge recall process within the model. We utilize the Multi-COUNTERFACT dataset (Meng et al. 2022b), in which each editing sample includes an input prompt, the corresponding correct answer and incorrect answer as shown in Fig. \ref{COUNTERFACT}.
We use the correct answer as the model's original answer and the incorrect answer as the target editing answer, the same as the original usage of this dataset.
The incorrect answer is related to the correct answer, satisfying the relation. By comparing the information changes in the representations of these two answers, our exploratory experiments not only corroborate previous findings on model interpretability but also uncover a critical process for knowledge editing: Probability Promotion. However, in the higher layers, the probability of the final output at the prediction position does not consistently increase, instead, it decreases in the topmost layers. To investigate the mechanisms behind this prediction elimination phenomenon, we extend our probing method to additional categories of information.

\subsection{Probe on Related Information}
In addition to the original correct answers, we construct word sets representing different information related to each input prompt (\textless subject, relation, object\textgreater). Below we elaborate the process of formulating these sets. \textbf{1. Related Set}. Firstly, we collect all the words related to the current factual statement. Using the subject and relation as query, we extract the top 100 relevant paragraphs from English Wikipedia by BM25, keeping paragraphs where the subject appears in either the title or content (Geva et al. 2023). After dedeuplicating and removing the object, we get a word set $\mathcal{S}$ related to the specific factual statement. \textbf{2. Factual-Related Set}. Considering that not all retrieved words pertain to factual information, we distinguish between factual and non-factual words by using GPT-4. For the subject, we ask GPT-4 to generate factual information related to it, including potential relations and their corresponding objects. Similarly, for each relation, GPT-4 is tasked with producing lists of subjects and objects sharing the specified relation. GPT-4 generates 50 candidates for each subject or relation. The resulting deduplicated collection forms the factual-related word set $\mathcal{S}_{factual}$. \textbf{3. Stopwords Set}. We observe that stopwords are indispensable regardless of whether a sentence related to factual information. Therefore, we construct $\mathcal{S}_{stop}$, comprising various non-informative words such as punctuation, conjunctions, prepositions, and others. \textbf{4. Non-Factual-Related Set}. Excluding $\mathcal{S}_{factual}$ and $\mathcal{S}_{stop}$ from $\mathcal{S}$ results in the non-factual word set $\mathcal{S}_{non-factual}$. . Notably, while there is no overlap in the words among the $\mathcal{S}_{factual}$, $\mathcal{S}_{non-factual}$ and $\mathcal{S}_{stop}$, tokenization may introduce overlapping tokens, which does not substantially alter the general trend of information variation.

\begin{figure}[t]
\centering
\includegraphics[scale = 0.275]{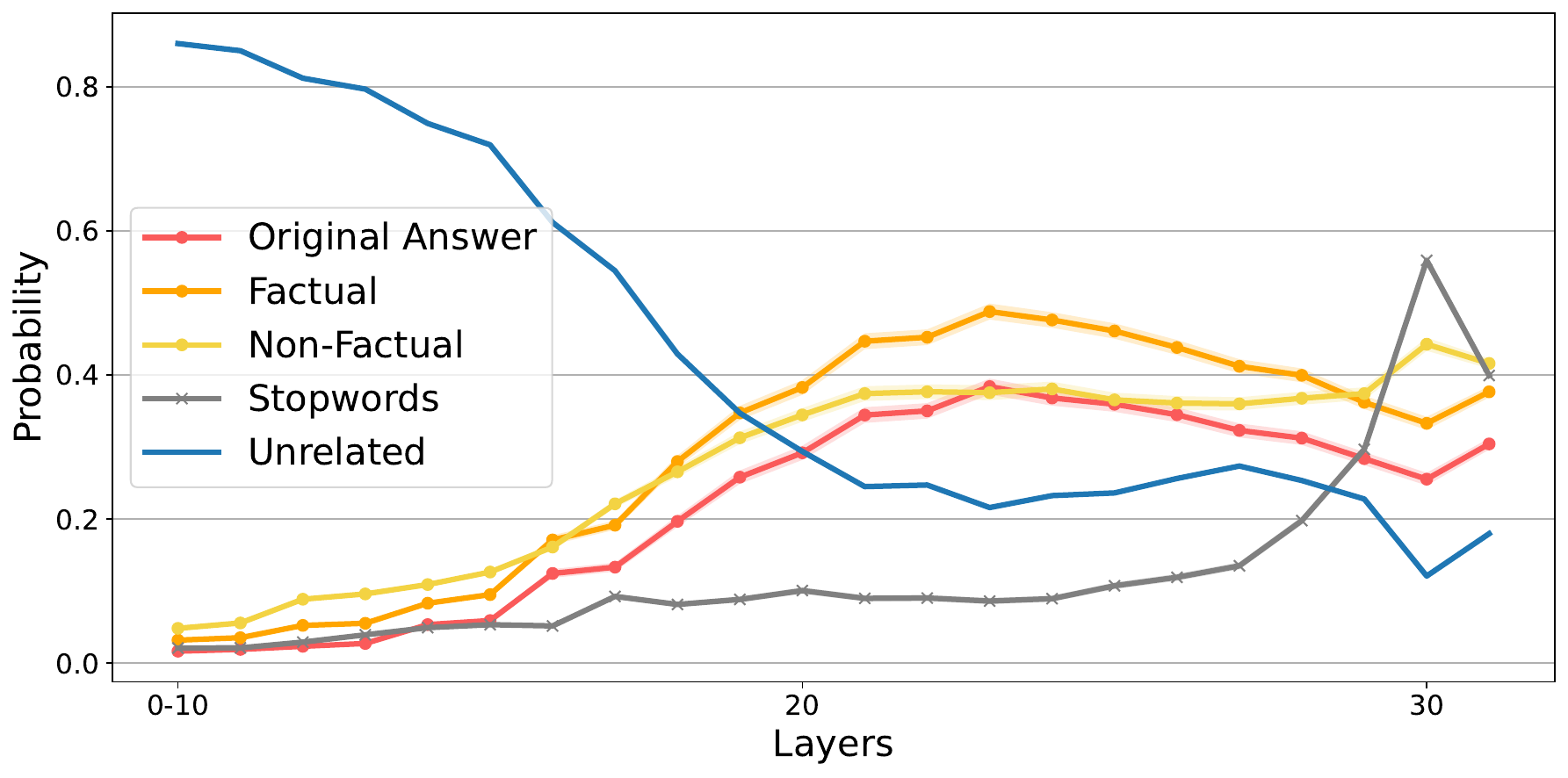}
\caption{Information change of different sets at the last prediction position, indicated by their average probability in the representations.} 
\label{full_info}
\end{figure}

As in \S 4.1, we input the 10,000 factual prompts to LLaMA (7B) and record the information changes of these sets at the last prediction position in Fig. \ref{full_info}, indicated by their average probability in the representations. The trajectory of factual information mirrors that of the original answers. As their information decreases in higher layers, non-factual elements, notably stopwords, receive enhanced promotion. The probability of stopwords significantly increases precisely when the probability of the original answer decreases, suggesting that the model's higher layers normalize the overall distribution. Considering that all input prompts contain stopwords, it can be inferred that the model aligns outputs with a realistic data distribution in the top-most layers. This finding does not conflict with our previous conclusions and does not affect the final prediction of the  original answer. Therefore, we did not update the model's highest layers during knowledge editing.

\begin{figure*}[t]
\centering
\includegraphics[scale = 0.5]{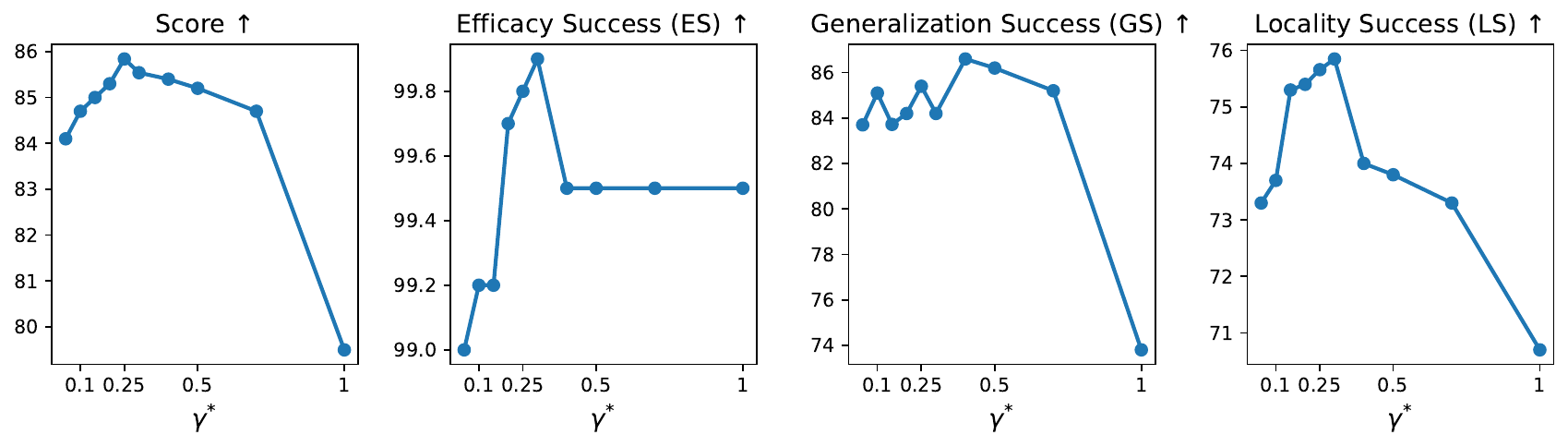}
\caption{Performances change of different clamp ratio $\gamma'$ for the high-layers updates.} 
\label{clamp}
\end{figure*}

\subsection{Probe on Unrelated Information}
Apart from the related information, we also investigate the change  of unrelated information. We apply the same probe approach to examine the variations in information content within the left token set $\mathcal{S}_{unrelated}$. This set is derived by excluding the aforementioned four word sets and the original correct answer from the entire vocabulary, followed by tokenization and deduplication. In Fig. \ref{full_info}, initially, a considerable amount of non-relevant information is present, as anticipated. With the progression to higher layers, this information continually diminishes. Intriguingly, in the high layers, such non-relevant information experiences a resurgence. This observation corroborates our hypothesis that the model’s higher layers function to regulate and balance the overall information. However, this resurgence of irrelevant information is not as pronounced as that of stopwords at higher levels, implying there is a focus in the model's smoothing mechanism. Given that stopwords are ubiquitous in all texts, the model's tendency to smooth the distribution at higher levels seems aimed at aligning the output more closely with an realistic information distribution.


\section{Analysis of Joint Updates}\label{APP: analysis}
In Section 4.3, we argue that changes in the low layers' information enrichment  should be greater than the changes in the high layers' probability promotion, meaning the clamp ratio $\gamma$ corresponding to the lower layers should be larger then $\gamma'$ to the high layers. To validate our hypothesis, we conduct an ablation study, keeping $\gamma$=1 constant while varying the $\gamma'$ from 0.05 to 1 to adjust the extent of updates in the high layers. Table \ref{clamp} presents the performances of simultaneously editing 10,000 samples from Multi-COUNTERFACT on the LLaMA (7B) model. It can be observed that the high-performing values of $\gamma'$  significantly smaller than $\gamma$=1, indicating that the required changes in the high layers are indeed minimal. Specifically, each metric initially increases and then decreases as $\gamma'$ grows. Since changes in the high layers are effective for learning new knowledge, if $\gamma'$ is too small, the $\delta'$ clamp operation after each update step will interfere with the $\delta'$ update process, leading to editing failure. Moreover, when $\gamma'$ becomes too large, the encoding of new knowledge is predominantly captured by $\delta'$. However, probability promotion requires minimal changes, and without the enrichment of information in the lower layers, relying solely on higher-layer modifications cannot achieve effective editing. Therefore, while the extent of updates in the higher layers should be smaller than in the lower layers, it must also be kept within an appropriate range.

\section{Benchmark Datasets}\label{APP: dataset}
In this section, we provide detailed descriptions, examples and evaluation metrics of the two benchmark datasets we use.
\begin{figure}[t]
\centering
\includegraphics[scale = 0.44]{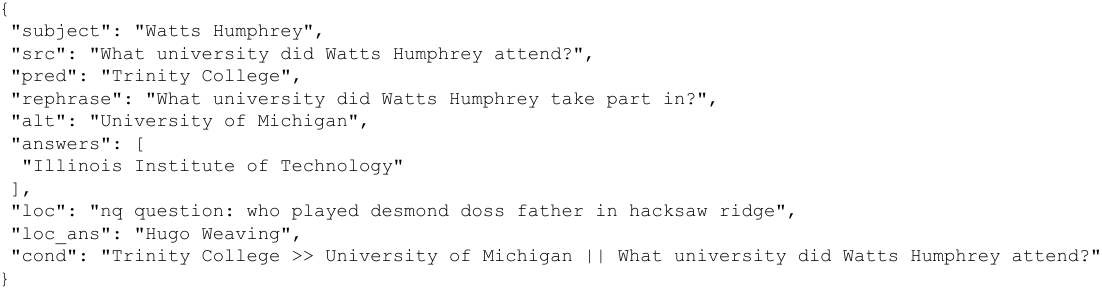}
\caption{An example of zsRE dataset.} 
\label{zsRE}
\end{figure}

\subsection{zsRE Dataset}
The information contained in every edit example of zsRE dataset is shown in Fig. \ref{zsRE}. Its editing target is the correct answer, results in a natural and realistic editing task to add factual knowledge.
The sample used for testing the generalization is the paraphrase of the input prompt. As for the locality, it uses unrelated facts about other subjects. Using the above samples, we test the editing method's Efficacy, Generalization and Locality by per-token success rate under the teacher-forcing method. Next we provide the formal definitions for the metrics. 

\textbf{Efficacy} measures the per-token accuracy on the edited samples. It is the proportion of cases where the model after editing could generate the correct answer's tokens on the same edited prompt:
\begin{equation}
\begin{aligned}
\mathbb{E}_{i}\left [ \mathbb{E}_{t\in \left \{ 1 \ldots T_i \right \}  }\left[ y_{i_t}^{\text{true}} = \underset{y_t}{\operatorname{argmax}} \mathbb{P}_{\mathcal{F}} \left[ y_t \mid x_i, y_{i_0}, \ldots, y_{i_{t-1}} \right] \right] \right ] . 
\label{}
\end{aligned}
\end{equation}

\textbf{Generalization} measures the success on the rephrasing prompts of the edited samples:
\begin{equation}
\begin{aligned}
\mathbb{E}_{i}\left [  \mathbb{E}_{x^{e} \in \textit{E}\mathrm (x_i)}\left[ \mathbb{E}_{t  }\left[ y_{i_t}^{\text{true}} = \underset{y_t}{\operatorname{argmax}} \mathbb{P}_{\mathcal{F}} \left[ y_t \mid x_i^e\ldots y_{i_{t-1}} \right] \right] \right] \right ].
\label{}
\end{aligned}
\end{equation}

\textbf{Locality} measures the success on the relevant facts of the same-type subject:
\begin{equation}
\begin{aligned}
\mathbb{E}_{i}\left [  \mathbb{E}_{x^{u} \in \textit{U}\mathrm (x_i)}\left[ \mathbb{E}_{t  }\left[ y_{i_t}^{\text{true}} = \underset{y_t}{\operatorname{argmax}} \mathbb{P}_{\mathcal{F}} \left[ y_t \mid x_i^u\ldots y_{i_{t-1}} \right] \right] \right] \right ]. 
\label{}
\end{aligned}
\end{equation}

\subsection{Multi-COUNTERFACT Dataset}
\begin{figure}[t]
\centering
\includegraphics[scale = 0.42]{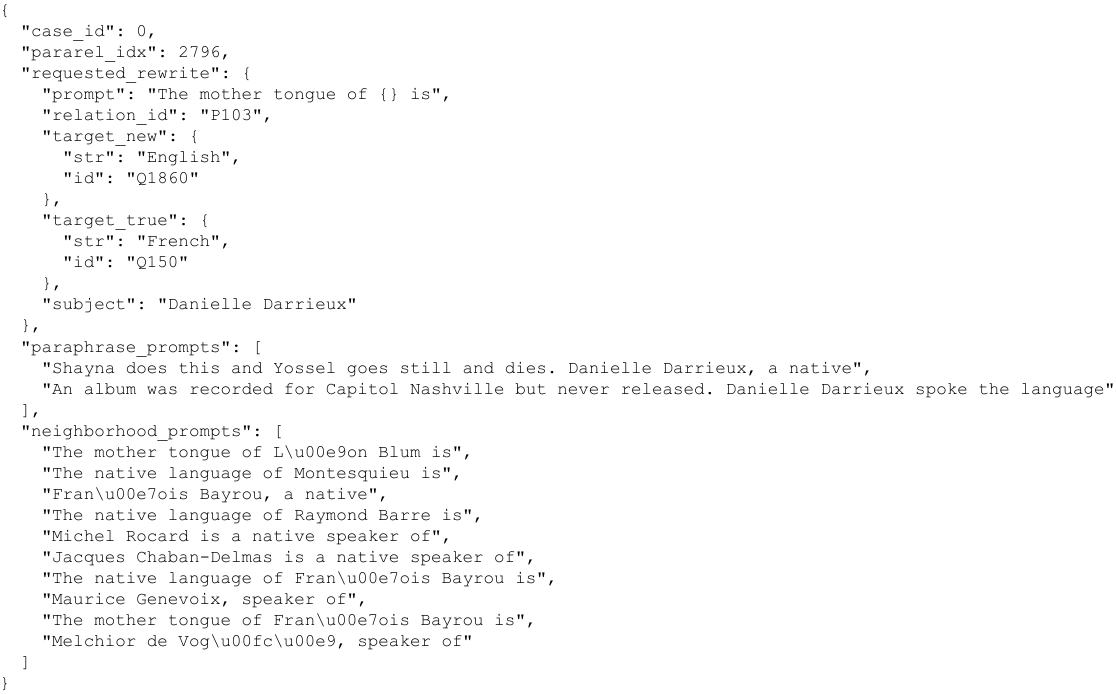}
\caption{An example of Multi-COUNTERFACT dataset.} 
\label{COUNTERFACT}
\end{figure}

The information contained in every edit example of Multi-COUNTERFACT dataset is shown in Fig. \ref{COUNTERFACT}. Each editing prompt is decomposed into subject, relation, true target, and new target, with the new target serving as the editing objective, making the editing task to inject counter-factual knowledge. Therefore, completing editing on this dataset is difficult. Moreover, it constructs paraphrases by both rephrasing the edit prompts and preceding an unrelated text.  As for the neighborhood prompt, it uses the facts of relevant subjects under the same relation. We detail the computation of each metric below. 

\textbf{Efficacy} measures the proportion of cases where the generation of the edited model having a larger probability for new target than true target on the same edited prompt:
\begin{equation}
\begin{aligned}
\mathbb{E}_{i}\left [ \mathbb{P} _{\mathcal{F}} \left [y_{i}^{new} \mid x_i\right ] >  \mathbb{P} _{\mathcal{F}} \left [y_{i}^{true}\mid x_i\right ] \right ].  
\label{}
\end{aligned}
\end{equation}

\textbf{Generalization} assesses the success rate of cases where the probability of generating new target is larger than that of the true target on the equivalent prompts:
\begin{equation}
\begin{aligned}
\mathbb{E}_{i}\left [ \mathbb{E}_{x^{e} \in \textit{Equivalent}\mathrm (x_i)}\left [  \mathbb{P} _{\mathcal{F}} \left [y^{new}\mid x^e\right ]>  \mathbb{P} _{\mathcal{F}} \left [y^{true}\mid x^e\right ] \right ]        \right ].
\label{}
\end{aligned}
\end{equation}

\textbf{Locality} evaluates whether the model could still maintain the original true answer on the facts of the relevant subject after the editing. It is the proportion of cases where the true answer having a larger probability than that of the new answer on the unrelated prompts.

\begin{equation}
\begin{aligned}
\mathbb{E}_{i}\left [ \mathbb{E}_{x^{u} \in \textit{Unrelated}\mathrm (x_i)}\left [  \mathbb{P} _{\mathcal{F}} \left [y^{true}\mid x^u\right ]>  \mathbb{P} _{\mathcal{F}} \left [y^{new}\mid x^u\right ] \right ]        \right ].
\label{}
\end{aligned}
\end{equation}

\section{Implementation Details}\label{APP: imple}
We conduct experiments for GPT-J and LLaMA on Multi-COUNTERFACT and zsRE datasets with GPU A100. We report the average results across 3 runs with different random seeds.

\subsection{Fine-Tuning with Weight Decay (FT-WD)}
For GPT-J, we use the same hyper-parameters in in (Meng et al. 2022b). \textbf{FT-WD} fine-tunes parameters in the 21th layer with a language model loss function. The fine-tuning process is limited to a maximum of 25 training steps, with a learning rate set at 5e-5. To mitigate overfitting, we implement a soft weight decay and an early stopping criterion, halting optimization when the loss falls below 1e-2. For LLaMA, we fine-tune layer 23 with the early stop condition that the loss is less than 1e-1.

The editing time for processing 10,000 samples from the COUNTERFACT dataset is approximately 0.48 hours on GPT-J and 0.36 hours on LLaMA.

\subsection{Model Editing Networks with Gradient Decomposition (MEND)}
We evaluate \textbf{MEND} only on GPT-J. We use the hyper-network trained by (Meng et al. 2022a) and the same hyper-parameters it uses. It takes 98.25 seconds to execute 10,000 updates on GPT-J.

\subsection{Rank-One Model Editing (Rome)}
\textbf{ROME} update 5th layer for GPT-J and 7th layer for LLaMA based on the probe experiments. The other settings adheres to the original paper's settings [Meng et al., 2022a]: a maximum of 20 training steps, a learning rate of 5e-1, a weight decay coefficient of 0.5, and a KL divergence coefficient of 0.0625. ROME samples different prefixes, 10 length-5 and 10 length-10 prefixes. Covariance statistics are computed in fp32 precision on 100,000 samples from Wikitext. Details are in (Meng et al. 2022a).

It task 12.29 hr for 10,000 edits on GPT-J and 37 hours for LLaMA.

\subsection{Massive Editing for Large Language Model via Meta Learning (MALMEN)} 
For GPT-J, we use the default hyper-parameters in \textbf{MALMEN} (Tan, Zhang, and Fu 2023) for editing on the zsRE dataset: updating the last 6 layers  with a rank of 1920, block of 2, epoch of 1, learning rate of 1e-6 and meta learning rate of 1e-5. 
See details in (Tan, Zhang, and Fu 2023). We keep the same hyper-parameters for LLaMA. It should be emphasized that in addition to the editing sample, MALMEN requires a training dataset containing editing, equivalent and unrelated samples to train the hyper-network. Since only the zsRE training dataset satisfies the conditions, we train the hyper-network on it for GPT-J and LLaMA, then conduct editing on the zsRE and Multi-COUNTERFACT editing set.

It takes about 10 hours to train the hyper-network and the editing time for 10,000 samples is 4.4 hours for GPT-J and 6 hours for LLaMA.

\subsection{Mass-Editing Memory in a Transformer (MEMIT)}
For the GPT-J model, we adhere to the optimal editing hyper-parameters as specified in the original \textbf{MEMIT} (Meng et al. 2022b). Specifically, \textbf{MEMIT} modifies the layers $\left\{3,4,5,6,7,8 \right \}$. For the optimization of $\delta$, we use a learning rate of 5e-1, a maximum of 25 update steps, a weight decay coefficient $\beta$ of 0.5, and a KL divergence coefficient of 0.0625. The L2 norm clamp ratio $\gamma$ is set to 0.75  to the original hidden state. Covariance statistics are collected using 100,000 samples from Wikitext in fp32 precision, with an adjustment factor $\lambda$ of 15,000. For the LLaMA model, based on the probe results from the original research, we select layers $\left\{5,6,7,8,9,10 \right \}$ for editing. Several editing settings are consistent with those for GPT-J: up to 25 training steps, a learning rate of 5e-1, a weight decay coefficient of 0.5, and a KL divergence coefficient of 0.0625. Covariance statistics are also computed in fp32 using 100,000 samples from Wikitext. However, for LLaMA, we use 0.9 for $\gamma$  and  4000 for $\lambda$, and we perform the editing in fp16 precision.

The editing process for 10,000 samples takes approximately 9 hours for GPT-J and 3.6 hours for LLaMA, owing to the fp16 precision used for LLaMA.

\subsection{Precise Model Editing (PMET)}
For GPT-J, we maintain the original hyper-parameters as uesd in (Li et al. 2024). Given that \textbf{PMET} is an enhancement of MEMIT, we only list the hyper-parameters that differ from MEMIT: a learning rate of 2e-1, a maximum of 30 update steps, a KL divergence coefficient of 1, and a $\lambda$ of 6000. For LLaMA, we also perform model editing in fp16 precision for the specified layers $\left\{5,6,7,8,9,10 \right \}$. For the optimization of $\delta$, we use the following hyper-parameters: a learning rate of 5e-1, a maximum of 30 update steps, a KL divergence coefficient of 1, and an L2 norm clamp ratio $\gamma$ of 0.9 (1 for zsRE). The $\lambda$ for adjusting covariance is set to 4000 (6000 for zsRE).

Since the optimization of $\delta$ and the updating strategy are similar to MEMIT, the time required to edit 10,000 samples is approximately the same as that for MEMIT.

\subsection{Joint Knowledge Editing for Information Enrichment and Probability Promotion (JEEP)}
\textbf{JEEP} performs model editing in fp16 precision for both models.
For GPT-J, we target layers $\left\{3,4,5,6,7,8 \right \}$. and layers $\left\{15,16 \right \}$ for editing. During the optimization of $\delta'$ and $\delta^{*}$, we use a learning rate of 5e-1, a maximum of 30 update steps, and weight decay coefficients $\beta'$ and $\beta^{*}$ both set to 0.5, along with a KL divergence coefficient of 0.0625. The L2 norm clamp ratios $\gamma'$ and $\gamma^{*}$ for $\delta'$ and $\delta^{*}$ are 0.75 and 0.25, respectively (0.75 and 0.8 for zsRE). Covariance statistics are collected using 100,000 samples from Wikitext in fp32 precision. The adjustment coefficients $\lambda'$ and $\lambda^{*}$ for the low and high layers' updates are 2000 and 2000, respectively (6000 and 2000 for zsRE).
For LLaMA, we target layers $\left\{5,6,7,8,9,10 \right \}$ and layers $\left\{22,23,24 \right \}$ for editing. The optimization process for $\delta'$ and $\delta^{*}$ is consistent with that of GPT-J. The L2 norm clamp ratios  $\gamma'$ and $\gamma^{*}$ are 1 and 0.25, respectively (1 and 0.7 for zsRE). The adjustment coefficients $\lambda'$ and $\lambda^{*}$ are 2500 and 1000, respectively (6000 and 3000 for zsRE).

Editing 10,000 samples from Multi-COUNTERFACT takes approximately 3.3 hours for GPT-J and 3.6 hours for LLaMA. Note that to ensure a fair comparison with the single-region update methods MEMIT and PMET, we update the same lower layers as they do. Although we have more regions to update, our faster optimization process results in a comparable overall time.

\end{document}